\documentclass[12pt]{article}
\usepackage{natbib,rotating,subcaption}
\usepackage{makecell} 
\usepackage{amsmath,natbib,mathtools,algorithm,amsthm}
\usepackage{amsfonts}

\usepackage{graphicx,colonequals}
\usepackage{caption}
\usepackage{subcaption,commath}
\usepackage{pbox}
\usepackage{enumitem}
\usepackage{pifont,lscape}
\usepackage{amsthm}
\usepackage{lipsum}
\usepackage{bm}
\usepackage{mathrsfs}
\usepackage{varwidth}
\usepackage{breqn}
\usepackage{hyperref}
\usepackage{footnote}
\usepackage{orcidlink}
\usepackage{bbm,xfrac}

\usepackage[noend]{algpseudocode}
\algnewcommand\algorithmicinit{\textbf{Initialize: }}

\newcommand{\zig}{\hat{z}_{ig}}

\newcommand{\vecmu}{\mbox{\boldmath$\mu$}}

\newcommand{\vecx}{\mathbf{x}}
\newcommand{\vecy}{\mathbf{y}}
\newcommand{\veca}{\mathbf{a}}
\newcommand{\vecb}{\mathbf{b}}
\newcommand{\vecc}{\mathbf{c}}
\newcommand{\vecd}{\mathbf{d}}

\newcommand{\vecX}{\mathbf{X}}

\newcommand{\veclam}{\mbox{\boldmath$\lambda$}}
\newcommand{\vecpi}{\mbox{\boldmath$\pi$}}

\newcommand{\vecvartheta}{\boldsymbol\vartheta}
\newcommand{\vecSigma}{\mathbf\Sigma}

\newcommand{\tr}{\,\mbox{tr}}
\newcommand{\vecA}{\mathbf{A}}

\newcommand{\deriv}{\mathrm{d}}

\newcommand{\invsig}{\vecSigma^{-1}}

\usepackage{authblk} 

\title{FastManly: An EM-Gradient Algorithm for Manly Mixture Models}

\author[1]{Katharine M. Clark\orcidlink{0000-0002-6162-2300}}
\author[2]{Paul D. McNicholas\orcidlink{0000-0002-2482-523X}}

\affil[1]{\small Department of Mathematics \& Statistics, Trent University, Ontario, Canada.}
\affil[2]{\small Department of Mathematics \& Statistics, McMaster University, Ontario, Canada.}

\date{} 
\begin{document}

\maketitle

\begin{abstract}
A faster implementation of mixtures of Manly transformations is proposed. This method, called FastManly, uses Newton's method for optimization in an EM gradient algorithm instead of Nelder-Mead in a traditional EM. A gradient and full Hessian are derived. Simulations show improved performance with noticeable speedups.  \\[-10pt]

\end{abstract}

\section{Introduction}

Model-based clustering utilizes finite mixture models to identify and group similar data points by assuming that the data originate from a mixture of component distributions. Each cluster usually corresponds to a component of the mixture, characterized by its own probability density function. The density of a finite mixture model can be expressed as
\begin{equation*}
	f(\mathbf{x} \mid \boldsymbol{\vartheta}) = \sum_{g=1}^{G} \pi_g f_g(\mathbf{x} \mid \boldsymbol{\theta}_g),
\end{equation*}
where $\boldsymbol{\vartheta} = \{\pi_1, \dots, \pi_G, \boldsymbol{\theta}_1, \dots, \boldsymbol{\theta}_G\}$, $\pi_g > 0$ represents the mixing proportion for the $g$th cluster, with the constraint that $\sum_{g=1}^G \pi_g = 1$, and $f_g(\mathbf{x} \mid \boldsymbol{\theta}_g)$ denotes the density of the $g$th component with parameters $\boldsymbol{\theta}_g$. Maximizing the likelihood of this model estimates the parameters and assigns data points to clusters based on the probability that each point belongs to a specific component.

Traditionally, Gaussian distributions were employed as the component densities in mixture models, but the field has evolved to incorporate a broader range of distributions \citep{mcnicholas16b}. These include multivariate skewed distributions for handling asymmetric data, discrete distributions for categorical data, and matrix-variate distributions for complex data structures. 

The Gaussian distribution, being symmetric, is clearly unsuitable for data with cluster skewness. To address this, \cite{azzalini85} developed the univariate skew-normal distribution, which includes a shape parameter, $\lambda$, and was first used in model-based clustering by \cite{lin07b}. The skew-normal distribution was extended to the multivariate case by \cite{azzalini96} and then used in model-based clustering by \cite{lin09}.  Similarly, the t-distribution was extended to the skewed data case by \cite{azzalini03} and first used in model-based clustering by \cite{pyne09}; however, a far more computatinally efficient representation of the skew-t distribution, based on a variance-mean mixture, was used by \cite{murray14b}. Furhtermore, other skewed distributions have been introduced into model-based clustering, including the multivariate shifted asymmetric Laplace \citep[SAL;][]{franczak14}, generalized hyperbolic \citep{browne15}, normal inverse Gaussian \citep{karlis09}, and variance-gamma distributions \citep{mcnicholas17}. 

Another approach to handle skewness is to transform the data to normality. \cite{lo12} apply the Box-Cox transformation. Similarly, \cite{zhu18} use the Manly transformation where, in the univariate case,
\begin{equation*}
	Y=\begin{cases}
		\frac{e^{\lambda X}-1}{\lambda}, & \lambda \neq 0;\\
		X,  & \lambda= 0.
	\end{cases}
\end{equation*}
In the multivariate case, $\veclam=(\lambda_1, \lambda_2, \dots, \lambda_p)$ is chosen such that the data become $p$-dimensional multivariate normal. The density of a finite mixture model with these transformed data is given by 
\begin{equation*}
	f(\vecx\mid\vecvartheta)=\sum_{g=1}^{G}\pi_g \phi(\mathcal{M}(\vecx\mid \veclam_g)\mid\vecmu_g, \bm\Sigma_g)\exp\{\veclam_g^{\top}\vecx\},
	\label{eq:manlymixture} 
\end{equation*}
where $\pi_g>0$ is the mixing proportion such that $\sum_{g=1}^G \pi_g =1$, $\veclam_g=(\lambda_{g1}, \dots \lambda_{gp})^{\top}$ is the transformation vector for the $g$th component and $$\mathcal{M}(\vecX\mid \veclam_g)=  \left\{\frac{e^{\lambda_{g1}X_1}-1}{\lambda_{g1}}, \dots, \frac{e^{\lambda_{gp}X_p}-1}{\lambda_{gp}}\right\}$$ is the transformed variable. 

\cite{zhu18} develop an EM algorithm to model mixtures with components derived from the Manly transformation. In the E-step, component memberships are estimated using
\begin{equation*}
	\zig^{(k+1)}=\frac{\hat{\pi}_g^{(k)}\phi\Big(\mathcal{M}(\vecx_i\mid \hat\veclam_{g}^{(k)})\mid \hat\vecmu_g^{(k)},\hat\vecSigma_g^{(k)}\Big)\exp\Big\{(\hat\veclam_{g}^{(k)})^\top\vecx_i\Big\}}{\sum_{h=1}^{G}\hat{\pi}_h^{(k)}\phi\Big(\mathcal{M}(\vecx_i\mid \hat\veclam_{h}^{(k)})\mid \hat\vecmu_h^{(k)},\hat\vecSigma_h^{(k)}\Big)\exp\Big\{(\hat\veclam_{h}^{(k)})^\top\vecx_i\Big\}}. 
\end{equation*}
In the M-step, while $\pi_g$, $\vecmu_g$, and $\vecSigma_g$ have closed-form estimators, $\veclam_{g}$ does not. Instead, updating $\veclam_{g}$ requires maximizing the following expression:
\begin{equation}\label{eq:opt}
	\sum_{i=1}^N z_{ig}\Big[\log\phi\Big(\mathcal{M}(\vecx_i\mid \veclam_{g})|\vecmu_g, \vecSigma_g\Big) + \veclam_g^\top\vecx_i\Big]
\end{equation}
with respect to $\veclam_g$. Afterward, the remaining parameters are updated according to:
\begin{equation*}
	\hat\pi_g^{(k+1)}= \frac{N_g^{(k+1)}}{N}, \qquad \hat\vecmu_g^{(k+1)}=\frac{1}{N_g^{(k+1)}}\sum_{i=1}^N \hat z_{ig}^{(k+1)}\mathcal{M}(\vecx_i\mid \hat\veclam_{g}^{(k+1)}),
\end{equation*}
\begin{equation*}
	\hat\vecSigma_g^{(k+1)}=\frac{1}{N_g^{(k+1)}}\sum_{i=1}^N \hat z_{ig}^{(k+1)}\Big(\mathcal{M}(\vecx_i\mid \hat\veclam_{g}^{(k+1)})-\hat\vecmu_g^{(k+1)}\Big)\Big(\mathcal{M}(\vecx_i\mid \hat\veclam_{g}^{(k+1)})-\hat\vecmu_g^{(k+1)}\Big)^\top,
\end{equation*}
where $	N_g^{(k+1)}=\sum_{i=1}^n \hat z_{ig}^{(k+1)}$. 

In their seminal paper, \cite{zhu18} treat $\vecmu_g$ and $\vecSigma_g$ as functions of $\veclam_{g}$ and use Nelder-Mead optimization \citep{nelder65} to update $\veclam_{g}$ within each M-step. Nelder-Mead optimization is a method that repeatedly evaluates an objective function at the vertices of a simplex and adjusts the simplex to move toward a local optimum. The key advantage of the Nelder-Mead approach is that it requires a neither gradient nor Hessian.  However, this comes at the cost of repeated evaluations of the transformed data, of $\vecmu_g$ and $\vecSigma_g$, and then of the objective function.

\section{Methodology}\label{sec:newton}

The proposed method uses Newton's method because when derivative information is available, Newton's method can converge much more rapidly than Nelder-Mead. In addition, using an EM gradient algorithm achieves convergence while only evaluating $\vecmu_g$ and $\vecSigma_g$ once per M-step.
In multivariate optimization, Newton's method starts with an initial value $\veclam^0_{g}$ and generates a sequence $\{\veclam_{g}^n\}$ which converges to the value of $\veclam_{g}$ that minimizes the objective function. Each $\{\veclam_{g}^n\}$ is generated according to the recursion:
\begin{equation*}
	\veclam_{g}^{n+1}= \veclam_{g}^{n}-\mathbf{H}^{-1}\mathbf{G},
\end{equation*}
where $\mathbf{H}$ is the Hessian matrix of the objective function evaluated at at $\veclam_g^{n}$ and $\mathbf{G} = \nabla \mathbf{f}(\veclam_{g}^n)$ is the gradient of the objective function evaluated at $\veclam_g^{n}$. Estimating $\veclam_{g}$ requires maximizing \eqref{eq:opt} with respect to $\veclam_g$.  This is equivalent to minimizing the objective function
\begin{equation*}
	\mathcal{O}=-2\sum_{i=1}^n z_{ig}\Big[\log\phi\Big(\mathcal{M}(\vecx_i\mid \hat\veclam_{g})|\vecmu_g, \vecSigma_g\Big) + \veclam_g^\top\vecx_i\Big] 
\end{equation*}
with respect to $\veclam_g$.  Newton's method can be used to perform this minimization.
\subsection{Gradient Vector}\label{sec:gradient}

Each $\veclam_{g}$ is optimized separately.  For simplicity, we drop the subscripts, $g$, i.e. $\veclam = \veclam_{g}$, $\vecmu = \vecmu_g$, $\vecSigma = \vecSigma_g$. The $j$th element of the gradient vector is the partial derivative of the objective function with respect to $\lambda_j$. It is given by:
\begin{equation}\label{eq:gradient}
	G_j =\sum_{i=1}^n z_{ig}\left[\tr\left(\invsig\frac{\deriv\vecSigma}{\deriv\lambda_j}\right) +2\veca_i^\top\invsig\vecb^{(j)}_i - \veca_i^\top\vecd^{(j)}\veca_i - 2x_{i,j}\right],
\end{equation}
where $x_{i,j}$ is the $j$th variable for the $i$th observation, and 
\begin{align*}
	\veca_i &= \vecy_i-\vecmu,\qquad
	\vecb^{(j)}_i = \frac{\partial\veca_i}{\partial\lambda_j},\\
	\frac{\partial\vecSigma}{\partial\lambda_j} &= \frac{1}{n_g} \sum_{i=1}^N z_{ig} \left(\vecb^{(j)}_i\veca_i^\top + \veca_i\vecb_i^{(j)\top}\right),\\
	\vecd^{(j)}  &= -\frac{\deriv\invsig}{\deriv\lambda_j}= \vecSigma^{-1}\frac{\deriv\vecSigma}{\deriv\lambda_j}\vecSigma^{-1}.
\end{align*}
A detailed derivation of the gradient vector is given in Appendix~\ref{app:gradient}.

\subsection{Hessian}\label{sec:hessian}
The Hessian matrix represents the second-order partial derivatives of the objective function with respect to each $\lambda_{j}$ and $\lambda_{k}$. Each entry in the $p\times p$ matrix is calculated as  
\begin{equation} \label{eq:hessian}
	\mathbf {H}_{j,k}={\frac {\partial ^{2}\mathcal{O}}{\partial \lambda_{j}\,\partial \lambda_{k}}}=\sum_{i=1}^n z_{ig} {\frac {\partial ^{2}\mathcal{O}_i}{\partial \lambda_{j}\,\partial \lambda_{k}}},
\end{equation}
with
\begin{align*}
{\frac {\partial ^{2}\mathcal{O}_i}{\partial \lambda_{j}\,\partial \lambda_{k}}} &= \tr\left\{-\vecd^{(k)}\frac{\partial \vecSigma}{\partial \lambda_j} + \invsig\frac{\partial^2\vecSigma}{\partial\lambda_j\partial\lambda_k} \right\}
	 + 2\vecb_i^{(k)\top}\invsig\vecb_i^{(j)} - 2\veca_i^\top \vecd^{(k)}\vecb_i^{(j)}\\ &\qquad-2\vecb_i^{(k)\top}\vecd^{(j)}\veca_i -\veca_i^\top\frac{\partial \vecd^{(j)}}{\partial \lambda_k}\veca_i + \mathbbm{1}_{j=k}\left\{2\veca_i^\top\invsig\vecc^{(j)}_i \right\},
\end{align*} 
 where $\mathbbm{1}_{j=k} =1$ is an indicator function taking the value 1 when $j=k$ and 0 otherwise, and
\begin{align*}
	\frac{\partial^2\vecSigma}{\partial \lambda_{j}\,\partial \lambda_{k}} &= \frac{1}{n_g} \sum_{i=1}^N z_{ig} \left(\vecb^{(j)}_i\vecb_i^{(k)\top} + \vecb_i^{(k)}\vecb_i^{(j)\top}+\mathbbm{1}_{j=k}\left\{\vecc^{(j)}_i \veca_i^{\top} + \veca_i\vecc_i^{(j)\top}\right\}\right),\\
	\frac{\partial \vecd^{(j)}}{\partial\lambda_k} &= -\vecd^{(k)}	\frac{\partial\vecSigma}{\partial\lambda_j}\invsig + \invsig\frac{\partial^2\vecSigma}{\partial \lambda_{j}\,\partial \lambda_{k}} \invsig -\vecd^{(j)}	\frac{\partial\vecSigma}{\partial\lambda_k}\invsig ,\\
	\vecc_i^{(j)} &= \frac{\partial\vecb_i^{(j)}}{\partial \lambda_j}.
\end{align*}
The Hessian is fully derived in Appendix~\ref{app:hessian}.

\subsection{EM Gradient Algorithm}\label{sec:modnewton}

Fitting a model with Newton's method may require a long sequence of estimates for $\veclam_{g}$, just for one iteration of the algorithm. Instead, \cite{lange95} justifies the use of one step of Newton's method in the M-step, thereby creating the EM gradient algorithm. \cite{lange95} argues that Newton's method converges quadratically, suggesting that a single iteration of Newton's method at each M-step should be sufficient to achieve convergence for an generalized EM algorithm. All that is required is that each M-step increases the expectation of the complete-data log-likelihood. Intuitively, this choice is justified because early estimates of $\veclam_{g}$ are updated using  early, and likely unreliable, estimates for $\zig$.

Using the gradient and Hessian from Sections~\ref{sec:gradient} and \ref{sec:hessian}, respectively, FastManly, an EM gradient algorithm for modelling mixtures with components derived from the Manly transformation, is described in Algorithm~\ref{alg:newmanly2}.

\begin{algorithm}
	\caption{FastManly}\label{alg:newmanly2}
	\begin{algorithmic}[1]
		\State \algorithmicinit $\hat{z}_{ig}$, $\hat\veclam_{g}$
		\Repeat
		\State update $\hat\veclam_{g}= \hat\veclam_{g}^{(\text{old})}-\mathbf{H}^{-1}\mathbf{G}$
		\Comment{$\mathbf{H}$ and $\mathbf{G}$ are functions of $\hat\veclam_{g}^{(\text{old})}$.} 
		\State update $N_g=\sum_{i=1}^N \hat z_{ig}$
		\State update $\hat\vecmu_g=\frac{1}{N_g}\sum_{i=1}^N \hat z_{ig}\mathcal{M}(\vecx_i\mid \hat\veclam_{g})$
		\State update $\hat\vecSigma_g=\frac{1}{N_g}\sum_{i=1}^N \hat z_{ig}\big(\mathcal{M}(\vecx_i\mid \hat\veclam_{g})-\hat\vecmu_g\big)\big(\mathcal{M}(\vecx_i\mid \hat\veclam_{g})-\hat\vecmu_g\big)^\top$
		\State update $\hat{z}_{ig}$
		\Until{convergence}
		\end{algorithmic}
\end{algorithm}

\section{Computational Advantages}
We can exploit sparsity, symmetry, and linear solves to maximize computational efficiency.

\subsection{Gradient}
Expanding \eqref{eq:gradient}, we get
\begin{equation*}
G_j =\sum_{i=1}^n z_{ig}\left[\tr\left(\invsig\frac{\deriv\vecSigma}{\deriv\lambda_j}\right) +2\veca_i^\top\invsig\vecb^{(j)}_i - \veca_i^\top\invsig\frac{\deriv\vecSigma}{\deriv\lambda_j}\invsig\veca_i - 2x_{i,j}\right].
\end{equation*}
Notice that 
\begin{align*}
	\frac{\partial\vecy}{\partial\lambda_j} &= \left(0,\ldots , \frac{\nu_j\exp(\nu_j)-\exp(\nu_j)+1}{\lambda_{j}^2}, \ldots 0\right)^\top
\end{align*}
is sparse, with an entry only in the $j$th position. The same is true for $\frac{\partial\vecmu}{\partial\lambda_j}$. Thus, we can treat each $\vecb^{(j)}_i$ as a scalar, denoted by $b_{i}^{(j)}$, and store the collection of $\{b_{i}^{(j)}\}_{i=1,\ldots,N}$ as a vector of length $N$. We stack the vectors $\mathbf{a}_1,\ldots,\mathbf{a}_N$ as rows of a matrix $\mathbf{A}$, allowing operations on all vectors simultaneously. Additionally, we can avoid calculating $\invsig$ by precomputing $\invsig\vecA$ and $\invsig\frac{\deriv\vecSigma}{\deriv\lambda_j}$ by using a Cholesky decomposition of $\vecSigma$ and a linear solver. 

\subsection{Hessian}
Expanding \eqref{eq:hessian}, we get
\begin{equation}\label{eq:hessian2}
\begin{split}
\mathbf {H}_{j,k}&=\sum_{i=1}^n z_{ig}\Bigg[ \tr\left\{-\invsig\frac{\partial \vecSigma}{\partial \lambda_k}\invsig\frac{\partial \vecSigma}{\partial \lambda_j} + \invsig\frac{\partial^2\vecSigma}{\partial\lambda_j\partial\lambda_k} \right\} + 2\vecb_i^{(k)\top}\invsig\vecb_i^{(j)}\Bigg.\\
&\qquad - 2\veca_i^\top \vecd^{(k)}\vecb_i^{(j)} -2\vecb_i^{(k)\top}\vecd^{(j)}\veca_i -\veca_i^\top\frac{\partial \vecd^{(j)}}{\partial \lambda_k}\veca_i + \mathbbm{1}_{j=k}\left\{2\veca_i^\top\invsig\vecc^{(j)}_i \right\}\Bigg].
\end{split}\end{equation} 
For 
\begin{equation*}
	\tr\left(-\invsig\frac{\partial \vecSigma}{\partial \lambda_k}\invsig\frac{\partial \vecSigma}{\partial \lambda_j} + \invsig\frac{\partial^2\vecSigma}{\partial\lambda_j\partial\lambda_k} \right),
\end{equation*}
we can precompute $\invsig\frac{\partial^2\vecSigma}{\partial\lambda_j\partial\lambda_k}$ with the linear solver. We can also access stored $\invsig\frac{\partial \vecSigma}{\partial \lambda_j}$ and $\invsig\frac{\partial \vecSigma}{\partial \lambda_k}$.  For the second term, we notice that
\begin{equation*}
	2\vecb_i^{(k)\top}\invsig\vecb_i^{(j)}  = 2b_i^{(k)}b_i^{(j)} \mathbf{e}_k \invsig \mathbf{e}_j,
\end{equation*}
where $\mathbf{e}_j$  and $\mathbf{e}_k$ are unit vectors. Thus, we can precompute $\invsig \mathbf{e}_j$ with a linear solver, which is reused for each $i$. For the next three terms --- i.e., the terms after the trace and before the indicator function in \eqref{eq:hessian2}) --- we can reuse $\invsig\vecA$ and $\invsig\frac{\deriv\vecSigma}{\deriv\lambda_j}$, noting that $\mathbf{A}^\top\invsig = (\invsig\vecA)^\top$ and $\frac{\deriv\vecSigma}{\deriv\lambda_j}\invsig=(\invsig\frac{\deriv\vecSigma}{\deriv\lambda_j})^\top$. All $\vecSigma$ inversions in the fifth term involve terms precomputed by linear solver. Finally, for the last term, recall that $\vecc^{(j)}_i$  is a scalar and $\invsig\mathbf{A}$ is pre-computed.

\section{Simulation Study}\label{sec:volatile}
In this section, we compare FastManly with ManlyMix in terms of speed and accuracy. We also consider a third, parallelized version of FastManly.

\subsection{Simulation Scheme}
ManlyMix is implemented in \textsf{R} \citep{R25} using the \texttt{ManlyMix} package \citep{ManlyMix}, which uses C as a backend. FastManly is written in C++ and called into \textsf{R} using the \texttt{Rcpp}  package \citep{eddelbuettel26}.  A parallelized version of FastManly, henceforth called FastManlyPar is also written in C++, where each $\veclam_{g}$ is updated concurrently.  

To compare the methods, we generate 100 datasets for 25 scenarios, each with different combinations of $n\in \{3, 6, 12, 24, 48\}$ and $p\in \{300, 600, 1200, 2400, 4800\}$. Each dataset is generated using the \texttt{ManlyMix} \textsf{R} package with  the following parameters:
\begin{center}
	$\begin{matrix*}[l]
		\vecpi=(1/3,1/3,1/3)&&\vspace{5pt}\\
		\veclam_1=(2.5,\ldots, 2.5)^\top, &\veclam_2=(2.45,\ldots, 2.45)^\top, &\veclam_3=(2.55,\ldots, 2.55)^\top.
	\end{matrix*}$ \end{center}
In addition, each entry of $\vecmu_1$,  $\vecmu_2$, and $\vecmu_3$ is sampled in $[15,25]$,  $[10,25]$, and $[15,30]$, respectively. Finally, $\vecSigma_g$ is randomly generated using the \texttt{clusterGeneration} package \citep{clustergeneration}.

\subsection{Method}

Each scenario is executed independently on a computing cluster. For each replicate, a dataset is generated and then clustered using ManlyMix, FastManly, and FastManlyPar. Each algorithm is allowed a maximum of 1000 iterations and is given a convergence threshold of $\varepsilon = 10^{-5}$. To mitigate potential ordering effects, the sequence in which the algorithms are run is randomized for each replicate. This process is repeated 100 times for each scenario. Prior to each scenario, the cluster was warmed up by running a preliminary experiment consisting of 20 replicates with $p=3$ and $n=300$. The results from these runs are discarded.

\subsection{Results}
For each clustering, the log-likelihood is recorded and time is measured using elapsed time from the \texttt{system.time} function. Elapsed time reflects the total runtime experienced by the user, including all computation performed in \textsf{R} and C/C++, as well as the interface overhead between the two environments.

\subsubsection{Timing}
\begin{figure}[!htb]
	\centering
	\includegraphics[width=5.26in]{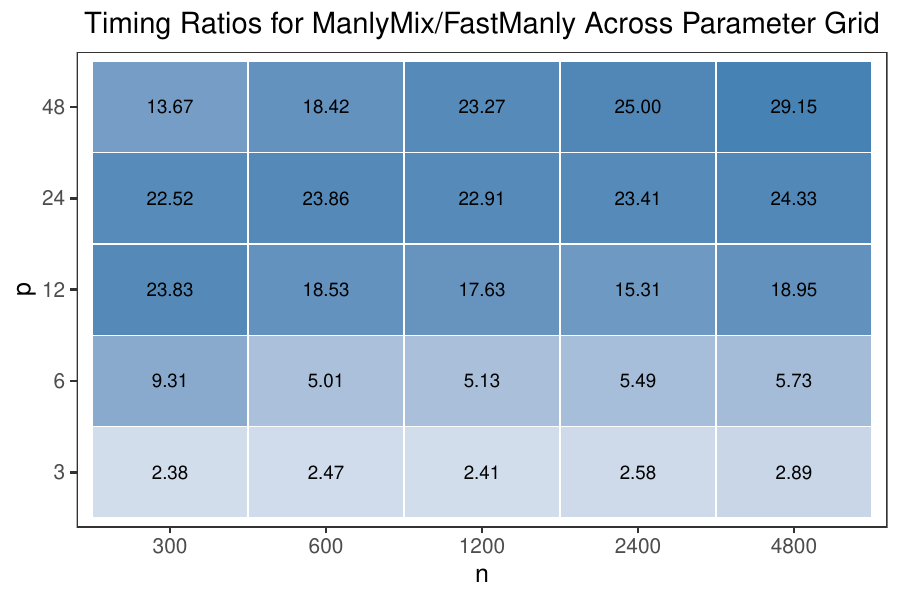}
	\caption{Geometric mean timing ratios, reflected as $\text{time}_{\text{ManlyMix}}/\text{time}_{\text{FastManly}}$. Cells are coloured according to $\log_2(\text{ratio})$}
	\label{fig:timinggrid}
\end{figure}

A grid is displayed in Figure~\ref{fig:timinggrid}, which displays the geometric mean ratio of time for ManlyMix and FastManly for each combination of $n$ and $p$. FastManly is at minimum 2.38 times faster than ManlyMix for low values of $n$ and $p$ and increases to 29.15 times at $p=48$, $n=4800$. While the ratio is favourable at all tested values, the absolute time difference is inconsequential at $p=3$, $n=300$, with ManlyMix taking on average 0.086 seconds and FastManly taking 0.037. However, the difference amplifies as $p$ and $n$ increase, with ManlyMix taking 603.77 seconds at $p=48$, $n=4800$, and FastManly only taking 20.95. This represents an average time difference of 9.71 minutes. Across all runs and scenarios, ManlyMix takes an average of 50.80 seconds, while FastManly takes 2.22 seconds. There is clearly a practical advantage to FastManly's speed.

\begin{figure}[!htb]
	\centering
	\includegraphics[width=5.26in]{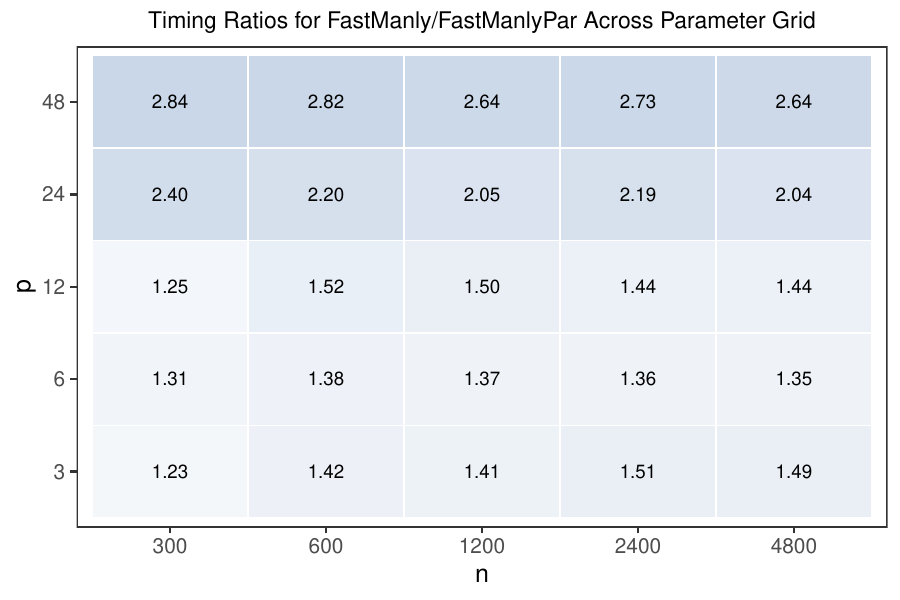}
	\caption{Geometric mean timing ratios, reflected as $\text{time}_{\text{FastManlyPar}}/\text{time}_{\text{FastManly}}$. Cells are coloured according to $\log_2(\text{ratio})$}
	\label{fig:timinggridpar}
\end{figure}

Figure~\ref{fig:timinggridpar} displays the timing ratios for FastManly compared to its parallelized version, FastManlyPar. Parallelizing the $\veclam_{g}$ updates across clusters provides a modest speedup, with this ratio reaching a maximum of 2.82$\times$. The advantage is likely capped shy of 3$\times$, because with 3 clusters, only three $\veclam_{g}$ updates can occur simultaneously. The E-step, however, cannot be parallelized. Just like with ManlyMix, the absolute time difference is inconsequential at $p=3$, $n=300$ (0.037 vs.\ 0.029 seconds), but the difference is more pronounced at $p=48$, $n=4800$ (20.95 vs.\ 7.92 seconds). 

\subsubsection{Clustering Performance}

While the speedup is beneficial, it is important that FastManly retain good likelihood maximization performance. As such, we compare the log-likelihoods from each method.  Unsurprisingly, FastManly and FastManlyPar achieve identical log-likelihoods. Therefore, we focus on the log-likelihoods from ManlyMix and FastManly. The differences in log-likelihood between FastManly and ManlyMix are displayed in Table~\ref{tab:loglik}, summarized by median difference, minimum difference in parentheses, and maximum difference in square brackets. 
\renewcommand{\arraystretch}{1.4}
\setcellgapes{3pt} \makegapedcells
 \begin{table}[ht] \centering
 	 \caption{Summary of median, (minimum), and [maximum] differences between FastManly's and ManlyMix's log-likelihoods for each scenario. Differences are taken FastManly--ManlyMix.}
 	 \label{tab:loglik}
 	 \begin{tabular}{|c|ccccc|} 
 	 	\hline 
 	 	&\multicolumn{5}{c|}{$p$}\\
 	 	\cline{2-6}
 	 	$n$ & 3 & 6 & 12 & 24 & 48 \\ 
 	 	\hline 	\hline
 	 	300 & \makecell{0.02\\(-3.78)\\{[23.04]}} & \makecell{0.00\\(-0.69)\\{[8.72]}} & \makecell{-0.00\\(-0.81)\\{[212.35]}} & \makecell{0.13\\(-2.52)\\{[323.00]}} & \makecell{6.28\\(-1.35)\\{[2993.53]}} \\
 	 	 \hline 
 	 	 600 & \makecell{0.09\\(-0.13)\\{[28.07]}} & \makecell{0.00\\(-0.36)\\{[144.84]}} & \makecell{0.00\\(-0.23)\\{[1.56]}} & \makecell{0.00\\(-0.65)\\{[22.55]}} & \makecell{0.03\\(-1.46)\\{[52.71]}} \\
 	 	 \hline
 	 	  1200 & \makecell{0.03\\(-4.33)\\{[10.14]}} & \makecell{0.00\\(-0.01)\\{[20.16]}} & \makecell{0.00\\(-0.16)\\{[0.09]}} & \makecell{0.00\\(-0.31)\\{[374.30]}} & \makecell{0.04\\(-0.51)\\{[35.41]}} \\
 	 	  \hline
 	 	   2400 & \makecell{0.12\\(-0.84)\\{[34.50]}} & \makecell{0.01\\(0.00)\\{[632.12]}} & \makecell{0.00\\(-0.01)\\{[0.02]}} & \makecell{0.00\\(-0.01)\\{[13.11]}} & \makecell{0.13\\(-0.00)\\{[143.51]}} \\ 
 	 	   \hline 
 	 	   4800 & \makecell{0.16\\(-8.67)\\{[161.79]}} & \makecell{0.01\\(0.00)\\{[1326.01]}} & \makecell{0.00\\(0.00)\\{[0.01]}} & \makecell{0.00\\(-0.01)\\{[0.05]}} & \makecell{0.23\\(0.01)\\{[194.33]}} \\ \hline 
	\end{tabular} 
	\end{table}

The median differences are all close to zero with the exception of $p=48$, $n=300$, indicating that each algorithm typically achieves a similar maximum log-likelihood. Minimum log-likelihood differences, in parentheses, indicate instances where ManlyMix achieves a higher log-likelihood than FastManly. Across all 2500 datasets, FastManly achieves a log-likelihood no worse than 8.67 below ManlyMix's. However, where FastManly achieves a larger log-likelihood, displayed in square brackets, the difference is sometimes two or three orders of magnitude above that. 

Not only does FastManly achieve similar maximization results, it often outperforms ManlyMix. This suggests that the EM gradient algorithm, with a single Newton step at each iteration, may provide a more effective optimization strategy. While the mechanism underlying this behaviour remains unclear, one possibility is that FastManly more effectively avoids local maxima, either through the use of different step sizes or by making $\zig$ updates between $\veclam_{g}$ updates.

\section{Conclusion}

FastManly is introduced as an alternative to the ManlyMix algorithm. Instead of optimizing $\veclam_{g}$ within each M-step using Nelder-Mead optimization, $\veclam_{g}$ is updated with a single step of Newton's method. The result is an EM gradient algorithm that outperforms ManlyMix in simulations in terms of elapsed time and likelihood maximization. A further parallelized version of FastManly provides additional gains in speed while achieving the same maximization performance as FastManly. 

\section*{Acknowledgements}
This work was funded by an NSERC Discovery Grant, a Dorothy Killam Fellowship, and the Canada Research Chairs program. 

\bibliographystyle{chicago}
\bibliography{masterbib.bib}

 \appendix
 \section{Derivations}
\subsection{Definitions}
Denote $\nu_{j} =\lambda_j X_{j}$ and 
\begin{equation}
	\vecy=\mathcal{M}(\vecX | \veclam)=\left(\frac{\exp(\nu_1)-1}{\lambda_{1}},\ldots , \frac{\exp(\nu_p)-1}{\lambda_{p}}\right)^\top.
\end{equation}
For convenience, define
\begin{align}
	\veca_i &= \vecy_i-\vecmu.
\end{align}
Further define
\begin{equation}
	\vecb^{(j)}_i = \frac{\partial\veca_i}{\partial\lambda_j}=  	\frac{\partial\vecy}{\partial\lambda_j} - \frac{1}{n_g} \sum_{i=1}^N z_{ig} \frac{\partial\vecy_i}{\partial\lambda_j}, 
\end{equation}
and equivalently $\vecb^{(k)}_i = \frac{\deriv\veca_i}{\deriv\lambda_k} $. Now, let
\begin{equation}
	\vecc^{(j)}_i= \frac{\partial\vecb_i}{\partial\lambda_j} = 	\frac{\partial^2\vecy}{\partial\lambda^2_j} - \frac{1}{n_g} \sum_{i=1}^N z_{ig} \frac{\partial^2\vecy_i}{\partial\lambda^2_j}
\end{equation}
and equivalently $ \vecc^{(k)}_i= \frac{\deriv\vecb_i}{\deriv\lambda_k}$. Finally define
\begin{equation}
	\vecd^{(j)}  = -\frac{\deriv\invsig}{\deriv\lambda_j}= \vecSigma^{-1}\frac{\deriv\vecSigma}{\deriv\lambda_j}\vecSigma^{-1}.
\end{equation} and equivalently $\vecd^{(k)}  = -\frac{\deriv\invsig}{\deriv\lambda_k}$.
The following quantities are further required:
\begin{align}
	\frac{\partial\vecy}{\partial\lambda_j} &= \left(0,\ldots , \frac{\nu_j\exp(\nu_j)-\exp(\nu_j)+1}{\lambda_{j}^2}, \ldots 0\right)^\top,\\
	\frac{\partial^2\vecy_i}{\partial\lambda^2_j}&= \left(0,\ldots , \frac{\left(\nu_j^2 - 2\nu_j + 2\right)  \exp\{\nu_j\}- 2}{\lambda_j^{3}}, \ldots 0\right)^\top.
\end{align}
Now define
\begin{equation}
	\vecmu = \frac{1}{n_g} \sum_{i=1}^N z_{ig} \vecy_i, 
\end{equation} 	and
\begin{equation}
	\vecSigma = \frac{1}{n_g} \sum_{i=1}^N z_{ig} (\vecy_i - \vecmu)
	(\vecy_i - \vecmu)^\top =  \frac{1}{n_g} \sum_{i=1}^N z_{ig} \veca_i
	\veca_i^\top. 
\end{equation}
Taking their derivatives with respect to $\lambda_j$ yields
\begin{equation}
	\frac{\partial\vecmu}{\partial \lambda_j} = \frac{1}{n_g} \sum_{i=1}^N z_{ig} \frac{\partial\vecy_i}{\partial\lambda_j} \quad\text{and}\quad 	\frac{\partial\vecSigma}{\partial\lambda_j} = \frac{1}{n_g} \sum_{i=1}^N z_{ig} (\vecb_i\veca_i^\top + \veca_i\vecb_i^\top).
\end{equation}

\subsection{Gradient}\label{app:gradient}
We aim to calculate
\begin{align*}
	\mathbf{G}_j =\frac{\partial}{\partial \lambda_j} \mathcal{O}
	&= \sum_{i=1}^n z_{ig} \frac{\partial}{\partial \lambda_j} \Big[-2\log\phi\Big(\mathcal{M}(\vecx_i\mid \hat\veclam)|\vecmu, \vecSigma\Big) -2 \veclam^\top\vecx_i\Big] \\
	&= \sum_{i=1}^n z_{ig} \frac{\partial}{\partial \lambda_j} \mathcal{O}_i,
\end{align*}
where 
	$\mathcal{O}_i = p\log(2\pi) + \log\det(\vecSigma) +\veca_i^\top \vecSigma^{-1}\veca_i -2 \veclam^\top \vecx_i$.
%
Now, The first term of $\sfrac{\partial}{\partial \lambda_j} \mathcal{O}_i$ is zero and the fourth term is $-2x_{i,j}$, where $x_{i,j}$ is the $j$th variable for the $i$th observation. The second term is 
\begin{equation*}
	\frac{\partial}{\partial \lambda_j}\log\det(\vecSigma)  = \tr\left\{\invsig\frac{\deriv\vecSigma}{\deriv\lambda_j}\right\},
\end{equation*}
while the third term is
\begin{equation*}
	\frac{\partial}{\partial \lambda_j} \veca_i^\top \vecSigma^{-1}\veca_i =2\veca_i^\top\invsig\vecb^{(j)}_i - \veca_i^\top\vecd^{(j)}\veca_i.
\end{equation*}
Putting this together, we get 
\begin{align*}
	\frac{\partial}{\partial \lambda_j} \mathcal{O}_i & = \tr\left\{\invsig\frac{\deriv\vecSigma}{\deriv\lambda_j}\right\} +2\veca_i^\top\invsig\vecb^{(j)}_i - \veca_i^\top\vecd^{(j)}\veca_i - 2x_{i,j},
\end{align*}
and 
\begin{equation*}
	\mathbf{G}_j =\sum_{i=1}^n z_{ig}\left[\tr\left\{\invsig\frac{\deriv\vecSigma}{\deriv\lambda_j}\right\} +2\veca_i^\top\invsig\vecb^{(j)}_i - \veca_i^\top\vecd^{(j)}\veca_i - 2x_{i,j}\right]
\end{equation*}

\subsection{Hessian}\label{app:hessian}
The Hessian matrix represents the second-order partial derivatives of the objective function with respect to each $\lambda_{k}$. Each entry in the $p\times p$ matrix is calculated as  
\begin{equation}
	\mathbf {H}_{j,k}={\frac {\partial ^{2}\mathcal{O}}{\partial \lambda_{j}\,\partial \lambda_{k}}}.
\end{equation}

\subsubsection{Main Diagonal}\label{app:maindiag}

We now take the partial derivative with respect to $\lambda_j$ for each term in \eqref{eq:gradient}. The partial derivative of first term is
\begin{equation*}
	\frac{\partial}{\partial \lambda_j}\tr\left\{\invsig\frac{\deriv\vecSigma}{\deriv\lambda_j}\right\} = \tr\left\{-\vecd^{(j)}\frac{\deriv\vecSigma}{\deriv\lambda_j} + \invsig\frac{\partial^2\vecSigma}{\partial\lambda_j^2}\right\},
\end{equation*}
where
\begin{equation*}
	\frac{\partial^2\vecSigma}{\partial \lambda_{j}^2} = \frac{1}{n_g} \sum_{i=1}^N z_{ig} \left(2\vecb^{(j)}_i\vecb_i^{(j)\top} + \vecc^{(j)}_i \veca_i^{\top} + \veca_i\vecc_i^{(j)\top}\right).
\end{equation*}
The derivative of second term is
\begin{equation*}
	\frac{\partial}{\partial \lambda_j}2\veca_i^\top\invsig\vecb^{(j)}_i = 2\vecb_i^{(j)\top}\invsig\vecb^{(j)}_i - 2\veca_i^\top\vecd^{(j)}\vecb^{(j)}_i + 2\veca_i^\top\invsig\vecc^{(j)}_i, 
\end{equation*} the derivative of third term is
\begin{equation*}
	\frac{\partial}{\partial \lambda_j}\{-\veca_i^\top\vecd^{(j)}\veca_i\}= -2\vecb^{(j)\top}_i\vecd^{(j)}\veca_i - \veca_i^\top\frac{\partial \vecd^{(j)}}{\partial\lambda_j}\veca_i,
\end{equation*}
where 
\begin{equation*}
	\frac{\partial \vecd^{(j)}}{\partial\lambda_j} = -2\vecd^{(j)}	\frac{\partial\vecSigma}{\partial\lambda_j}\invsig + \invsig\frac{\partial^2\vecSigma}{\partial\lambda_j^2}\invsig,
\end{equation*}
and fourth term's derivative is zero. Combining these, we get:
\begin{equation*}
	{\frac {\partial ^{2}\mathcal{O}_i}{\partial \lambda_{j}^2}} =  \tr\left\{-\vecd^{(j)}\frac{\deriv\vecSigma}{\deriv\lambda_j} + \invsig\frac{\partial^2\vecSigma}{\partial\lambda_j^2}\right\}+ 2\vecb_i^{(j)\top}\invsig\vecb^{(j)}_i - 4\veca_i^\top\vecd^{(j)}\vecb^{(j)}_i + 2\veca_i^\top\invsig\vecc^{(j)}_i - \veca_i^\top\frac{\partial \vecd^{(j)}}{\partial\lambda_j}\veca_i.
\end{equation*}

\subsubsection{Off-Diagonal}\label{app:offdiag}
We now take the partial derivative of \eqref{eq:gradient} with respect to $\lambda_k$, i.e.,
\begin{equation*}
	{\frac {\partial ^{2}\mathcal{O}_i}{\partial \lambda_{j}\,\partial \lambda_{k}}} =  {\frac {\partial }{\partial \lambda_{k}}}\left( \tr\left\{\invsig\frac{\deriv\vecSigma}{\deriv\lambda_j}\right\} +2\veca_i^\top\invsig\vecb^{(j)}_i - \veca_i^\top\vecd^{(j)}\veca_i - 2x_{i,j}\right),
\end{equation*} for $j\neq k$. For the first term, we calculate
\begin{equation*}
	\frac{\partial}{\partial \lambda_k}\tr\left\{\invsig\frac{\deriv\vecSigma}{\deriv\lambda_j}\right\} = \tr\left\{-\vecd^{(k)}\frac{\partial \vecSigma}{\partial \lambda_j} + \invsig\frac{\partial^2\vecSigma}{\partial\lambda_j\partial\lambda_k} \right\},
\end{equation*}
where
\begin{equation*}
	\frac{\partial^2\vecSigma}{\partial \lambda_{j}\,\partial \lambda_{k}} = \frac{1}{n_g} \sum_{i=1}^N z_{ig} \left(\vecb^{(j)}_i\vecb_i^{(k)\top} + \vecb_i^{(k)}\vecb_i^{(j)\top}\right), 
\end{equation*}
for $j\neq k$. The derivative of the second term is
\begin{equation*}
	\frac{\partial}{\partial \lambda_k}2\veca_i^\top\invsig\vecb^{(j)}_i \Big|_{j\neq k} = 2\vecb_i^{(k)\top}\invsig\vecb_i^{(j)} - 2\veca_i^\top \vecd^{(k)}\vecb_i^{(j)},
\end{equation*}
for $j\neq k$, noting that $\sfrac{\partial\vecb_i^{(j)}}{\partial \lambda_k} = 0$, when $j\neq k$. 
For the third term, 
\begin{equation*}
	\frac{\partial}{\partial \lambda_k}\left(-\veca_i^\top\vecd^{(j)}\veca_i\right) = -2\vecb_i^{(k)\top}\vecd^{(j)}\veca_i -\veca_i^\top\frac{\partial \vecd^{(j)}}{\partial \lambda_k}\veca_i,
\end{equation*} 
where 
\begin{equation*}
	\frac{\partial \vecd^{(j)}}{\partial\lambda_k} = -\vecd^{(k)}	\frac{\partial\vecSigma}{\partial\lambda_j}\invsig + \invsig\frac{\partial^2\vecSigma}{\partial \lambda_{j}\,\partial \lambda_{k}} \invsig -\vecd^{(j)}	\frac{\partial\vecSigma}{\partial\lambda_k}\invsig ,
\end{equation*}
and the fourth term is zero. Combining these, we get 
\begin{align*}
	\nonumber{\frac {\partial ^{2}\mathcal{O}_i}{\partial \lambda_{j}\,\partial \lambda_{k}}} \Big|_{j\neq k}&= \tr\left\{-\vecd^{(k)}\frac{\partial \vecSigma}{\partial \lambda_j} + \invsig\frac{\partial^2\vecSigma}{\partial\lambda_j\partial\lambda_k} \right\} + 2\vecb_i^{(k)\top}\invsig\vecb_i^{(j)} - 2\veca_i^\top \vecd^{(k)}\vecb_i^{(j)} \\
	&\qquad -2\vecb_i^{(k)\top}\vecd^{(j)}\veca_i -\veca_i^\top\frac{\partial \vecd^{(j)}}{\partial \lambda_k}\veca_i.
\end{align*} 

\subsubsection{Complete Hessian}
Thus, the complete Hessian is 
\begin{align*}
	{\frac {\partial ^{2}\mathcal{O}_i}{\partial \lambda_{j}\,\partial \lambda_{k}}} &= \tr\left\{-\vecd^{(k)}\frac{\partial \vecSigma}{\partial \lambda_j} + \invsig\frac{\partial^2\vecSigma}{\partial\lambda_j\partial\lambda_k} \right\} + 2\vecb_i^{(k)\top}\invsig\vecb_i^{(j)} - 2\veca_i^\top \vecd^{(k)}\vecb_i^{(j)}\\
	&\qquad -2\vecb_i^{(k)\top}\vecd^{(j)}\veca_i -\veca_i^\top\frac{\partial \vecd^{(j)}}{\partial \lambda_k}\veca_i
	 + \mathbbm{1}_{j=k}\left\{2\veca_i^\top\invsig\vecc^{(j)}_i \right\},
\end{align*} 
and 
\begin{equation}
	\mathbf {H}_{j,k} =\sum_{i=1}^n z_{ig} {\frac {\partial ^{2}\mathcal{O}_i}{\partial \lambda_{j}\,\partial \lambda_{k}}}.
\end{equation}

\end{document}